\documentclass[runningheads]{llncs}
\usepackage[pdftex]{graphicx}
\usepackage{xcolor}
\usepackage{multirow}
\usepackage{epstopdf}
\usepackage{xcolor}
\usepackage{multirow}
\usepackage[pagebackref=true,breaklinks=true,letterpaper=true,colorlinks,bookmarks=false]{hyperref}
\graphicspath{{pdf/}{jpeg/}}
\DeclareGraphicsExtensions{.pdf,.jpeg,.png}
\usepackage[cmex10]{amsmath}
\usepackage[caption=false,font=normalsize,labelfont=sf,textfont=sf, position=top]{subfig}
\usepackage{graphicx}
\usepackage{amsmath,amssymb} % define this before the line numbering.
\usepackage{color}

\begin{document}
\pagestyle{headings}
\mainmatter

%===========================================================
\title{Object Boundary Guided Semantic Segmentation} % Replace with your title

\author{Qin Huang, Chunyang Xia, Wenchao Zheng, Yuhang Song, Hao Xu, C.-C. Jay Kuo}
\institute{(qinhuang@usc.edu)}

\maketitle

%===========================================================
\begin{abstract}
Semantic segmentation is critical to image content understanding and
object localization.  Recent development in fully-convolutional neural
network (FCN) has enabled accurate pixel-level labeling.  One issue in
previous works is that the FCN based method does not exploit the object boundary information
to delineate segmentation details since the object boundary label is
ignored in the network training. To tackle this problem, we introduce a
double branch fully convolutional neural network, which separates the
learning of the desirable semantic class labeling with mask-level object proposals guided by relabeled boundaries.  This network, called
object boundary guided FCN (OBG-FCN), is able to integrate the distinct
properties of object shape and class features elegantly in a fully
convolutional way with a designed masking architecture.
We conduct experiments on the PASCAL VOC segmentation benchmark, and show that
the end-to-end trainable OBG-FCN system offers great improvement in optimizing the
target semantic segmentation quality. 
\end{abstract}

%===========================================================

\section{Introduction}\label{sec:introduction}

The convolutional neural network (CNN) has brought a rapid progress in
computer vision research and development in recent years
\cite{krizhevsky2012imagenet,simonyan2014very,szegedy2015going}. Due
to the availability of a large amount of image data \cite{Everingham10,lin2014microsoft,deng2009imagenet}, the performance of various CNNs has
been improved significantly. These deep learning based approaches have been applied to high-level vision
challenges such as image recognition and object detection
\cite{krizhevsky2012imagenet,girshick2014rich,girshick2015fast,ren2015faster} and low-level vision problems such as semantic
segmentation \cite{long2015fully,chen2014semantic,zheng2015conditional}.  The network learns to design tailored feature
pools for a vision task by examining deep features of discriminative
properties and shallow features of local visual patterns. 

Recent developments in the fully convolutional neural network (FCN)
\cite{long2015fully} have extended CNN's capability from image-level
recognition to pixel-level decision. It allows the network to see the
object location as well as the object class.  By taking the advantage of
low-level pooling features and probability distributions of neighboring
contents, recent studies \cite{long2015fully,chen2014semantic,zheng2015conditional} have further improved segmentation accuracy on the
PASCAL VOC dataset \cite{Everingham10}. 

One way to refine segmentation results is to exploit the edge
information \cite{chen2015semantic,bertasius2015semantic}. The ground
truth labels provided by the PASCAL VOC dataset have already offered the object
contour information.  However, object boundaries and hard cases are
marked with the same label. To avoid confusion, both object boundaries
and hard cases are ignored during the loss calculation in the training
stage. 

\begin{figure}
\centering
\subfloat{\includegraphics[width=4.7in]{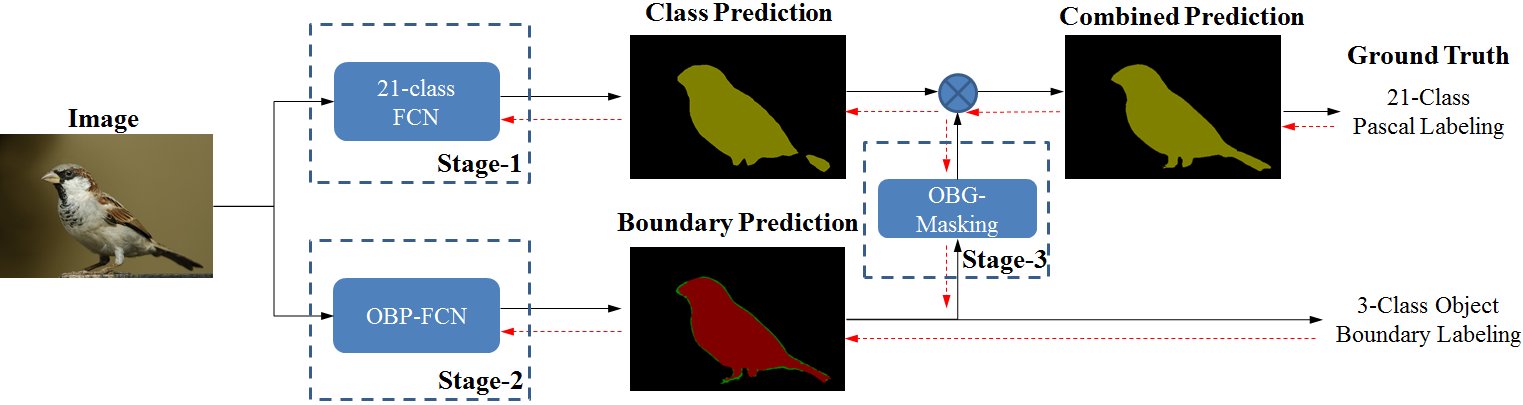}%
\label{p-1}} \hfil
\caption{The proposed fully convolutional OBG-FCN consists of three subnets: the FCN-8s, the
Object Boundary Prediction FCN (OBP-FCN) and the
Object Boundary Guided Mask (OBG-Mask).  In the first stage, the FCN-8s
is trained using the ground truth labels and used to predict object
classes. In the second stage, we convert the ground truth labels into 3
categories (object without class distinction, background and object
boundary) and train the OBP-FCN accordingly. In the third stage, we use
the OBG-Mask subnet to pass the boundary information to the result of
the FCN-8s to yield the ultimate semantic segmentation result.}\label{ex1}
\end{figure}

In this work, we propose an end-to-end fully convolutional neural network, which takes advantage of object boundaries to guide the
semantic segmentation. By relabeling the ground truth into three classes (object without class difference, object boundary and background),
we first independently train an object boundary prediction FCN (OBP-FCN), which gives us an accurate prior knowledge of object localizations and shape details.
This mask-level object proposal, then goes through a designed masking architecture (OBG-Mask), and is later combined with another FCN branch which specifically
learn to predict the object classes, formulating the object boundary guided FCN (OBG-FCN). The finalized system is thus able to combine the strengths of two independent pre-trained FCNs and
refine the output with the standard back-propagation, as illustrated in Fig. \ref{ex1}. 

%===========================================================
%===========================================================

We evaluate the performance of the proposed OBG-FCN method on the PASCAL
VOC 2011 and 2012 semantic segmentation datasets. It offers
great improvement compared with the baseline FCN model on both validation and testing sets.  The
experimental results demonstrate that object boundaries offer useful
information in delineating object details for better semantic
segmentation. 

The rest of this paper is organized as follows. The related work is
reviewed in Section \ref{sec:review}. The label conversion and the design
of the OBP-FCN are discussed in Section \ref{sec:OBP-FCN}. The full
OBG-FCN system is proposed in Section \ref{sec:OBG-FCN}. The
experimental results are presented in Section \ref{sec:exp}. Finally,
concluding remarks and future research directions are given in Section
\ref{sec:conclusion}.

\section{Related Work}\label{sec:review}
Being apart from the traditional segmentation task \cite{zhang1996survey,shi2000normalized,felzenszwalb2004efficient}, semantic
segmentation demands both pixel-wise accuracy and semantic
outputs. Thus, low-level image features and high-level
object knowledge have to be integrated to achieve this goal.  Deep
learning methods have been proposed and proven to be effective for
semantic segmentation. In this section, we review several related
work along this direction. 

Object detection is a topic that is highly related to semantic
segmentation. It has been extensively studied using CNNs, e.g.,
\cite{girshick2014rich,girshick2015fast,hariharan2014simultaneous,hariharan2015hypercolumns,he2015spatial}.  By
predicting object bounding boxes and categories, RCNN
\cite{girshick2014rich}, SPPnet \cite{he2015spatial} and Fast RCNN
\cite{girshick2015fast} can detect object regions using object
proposals.  Faster RCNN \cite{ren2015faster} exploits the shared
convolutional features to extract object proposals, leading to a faster
inference speed. Masking level proposals can also be extracted in a similar
manner by sharing either convolutional features or layer outputs
\cite{arbelaez2014multiscale,carreira2012cpmc,uijlings2013selective,dai2015convolutional}.  

The FCN \cite{long2015fully} allows pixel-wise regression.  By
leveraging the skip architecture \cite{bishop2001bishop} to combine the
information from pooling layers, the FCN can achieve coarse segmentation
with rough object boundaries.  MRF/CRF-driven CNN methods have been used
to train classifiers and graphical models simultaneously
\cite{russell2009associative,krahenbuhl2012efficient,chen2014semantic}
to further improve detection accuracy and segmentation details.  An
end-to-end framework has been proposed in \cite{zheng2015conditional} to
combine the conditional random field with the recursive neural network
(RNN) \cite{pinheiro2013recurrent} for performance enhancement to refine
 segmentation details. 

Recent developments in instance segmentation demonstrate the advantages
of multi-task learning and multi-network assembling.  For example, the
bounding box locations and object scores are predicted in a
fully-convolutional form in \cite{ren2015faster}. Furthermore, a
multi-task network cascades (MNCs) structure is proposed in
\cite{dai2016instance}.  This structure utilizes the result of a sub-task as a
pixel-level mask to help other subtasks in the network. The network
involves several subnetworks (or subnets) and considers their mutual
interaction to offer a powerful solution.  The network training can be
simplified by adopting an independent pre-training procedure for each
subnet which is then followed by a dependent learning procedure. 

One way particular in multi-tasking learning is to incorporate
edge/contour detection with semantic class labeling.  Specifically,
Bertasius et al. \cite{bertasius2015semantic} and Chen
\cite{chen2015semantic} exploit features from intermediate layers of a
deep network and conduct a edge detection sub-task in the similar way of
\cite{xie2015holistically}. In \cite{bertasius2015semantic}, Bertasius
et al. improves the boundary detection with semantic segmentation, while
Chen \cite{chen2015semantic} designs a domain transform structure to
conduct an edge-preserving filtering for segmentation. 

In comparison with previous related work, we propose a multi-network
system that addresses the object boundary detection and the semantic
segmentation problem simultaneously. It is shown that an improved
object boundary predictor can guide the object labeling task in semantic
segmentation.

\section{Object Boundary Prediction with OBP-FCN} \label{sec:OBP-FCN}

The FCN in \cite{long2015fully} is trained using the PASCAL VOC dataset
for the recognition of 20 object classes, which offers good
performance since it recognizes patterns of desired classes by examining
both coarse-level and fine-level visual features. It generates a blob-wise result to
describe the coarse shape of an object and predict its class label.
Although the deconvolution layer can partially recover the lost
resolution of the input in the pooling layer, its segmented result is
still rough and the class label could be wrong as local features can be
confusing. Edge detection is conducted in \cite{chen2015semantic} with
middle-level features, yet this method detects edges around and inside
an object. To enhance the accuracy of segmented object boundaries, we
propose a variant of the FCN, called the OBP-FCN, that offers pixel-wise
object/boundary prediction in this section. 

\subsection{Generation of New Labels} \label{labelob}

We first process the existing labels in the PASCAL-VOC dataset and convert
them into a set of new labels. Then, the new labels will be used to train
the OBP-FCN for more accurate object mask prediction. 

The PASCAL VOC dataset provides labels for object classes and instances
as the ground truth. For each image with indexes $I$,
$N$ object classes are labeled in $N$ colors, denoted by
$\L_{c}=\{l_{1},l_{2},\cdots,l_{N}\}$, where $N=20$ for the PASCAL VOC dataset.
The background area ($I_{b}$) is labeled in black, denoted by $l_{b}$, and 
region of the object boundary area and hard cases ($I_{w}$) are labeled in white, denoted by
$l_{w}$, which are usually ignored in the penalty function
calculation during the CNN training process.

To recover the accurate location information of object boundaries, we
convert the existing PASCAL VOC labels into our desired 4 categories: 1) objects
without class distinction ($l_o$), 2) object boundaries ($l_{ob}$), 3)
background ($l_b$), and 4) hard cases ($l_{hc}$). 

To begin with, we first derive the object indexes with labels $L_{I} \in \L_{c}$ as object regions ($I_{o}$).
We then derive the outline of each object region as the object boundary ($I_{ob}$).
For this purpose, we compare the label of each object pixel with those of
its neighbors in a $3 \times 3$ window, and label the one without uniform-class neighbor as object boundary. As a
result, we can find all pixels that separate different class labeling (object
classes as well as background). 

%%%%%%%%%%%%%%%%%%%%%%%%%%%%%%%%%%%%%%%%%%%%%%%%%%%%%%%%%%
\noindent
\noindent\rule[0.25\baselineskip]{\textwidth}{1.5pt}
{\it \textbf{Algorithm A} Label Conversion for the PASCAL VOC dataset.}\\
\noindent\rule[0.25\baselineskip]{\textwidth}{1pt}\\
\noindent
$ \triangleright$Initialize all image pixels with the background label, $\;\;\;\;\;\;\;\;\;\;\;\;\;\;\;\;\;\;\;\;\;\;\;\;\;\;\;\; L_I  \leftarrow l_{b} $\\
$ \triangleright$Assign object boundary label to extended region, $  \; \;\;\;\;\;\;\;\;\;\;\;\;\;\;\;\;\;\;\;\;\;\;\;\;\;\;  L_{I_{exb}} \leftarrow l_{ob}$\\
$ \triangleright$Assign object label to original object region, $	  \;\;\;\;\;\;\;\;\;\;\;\;\;\;\;\;\;\;\;\;\;\;\;\;\;\;\;\;\;\;\;\;\;\;\;\;\;\;\;\; L_{I_{o}} \leftarrow l_{o}$ \\
$ \triangleright$Label the hard case regions $\;\;\;\;\;\;\;\;\;\;\;\;\;\;\;\;\;\;\;\;\;\;\;\;\;\;\;\;\;\;\;\;\;\;\;\;\;\;\;\;\;\;\;\;\;\;\;\;\;\;\;\;\;\;\;\;\;\;\;\;\;\;\; L_{I_{hc}} \leftarrow l_{hc}  $\\
\noindent\rule[0.25\baselineskip]{\textwidth}{1pt}
%%%%%%%%%%%%%%%%%%%%%%%%%%%%%%%%%%%%%%%%%%%%%%%%%%%%%%%%%%

Then we thicken the boundary region ($I_{ob}$) into the extended
boundary region ($I_{exb}$) by dilating a pixel to four directions by $w$ 
pixels. The remaining pixels with original $l_{w}$ label and not within the thickened boundary region
are noted as hard cases ($I_{hc}$).

After deriving all desirable class regions, we assign the target labels
as in Algorithm A. Note that the order of the assignment is very important
so as to keep the completeness of object and the accuracy of the boundary.

\begin{figure}
\captionsetup[subfigure]{}
\centering
\subfloat[Image]{\includegraphics[width=1.4in]{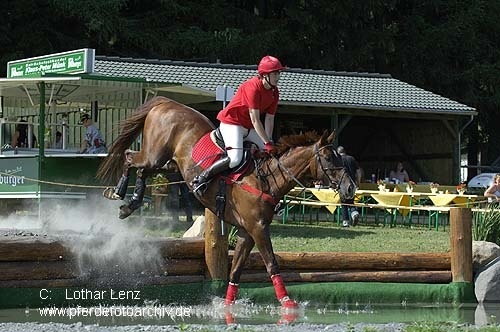}%
\label{p-1}} \hfil
\subfloat[Original Labels]{\includegraphics[width=1.4in]{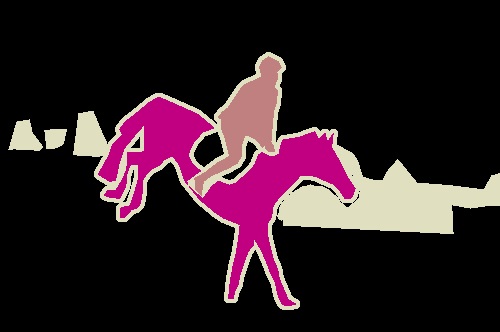}%
\label{p-2}}\hfil
\subfloat[New Labels]{\includegraphics[width=1.4in]{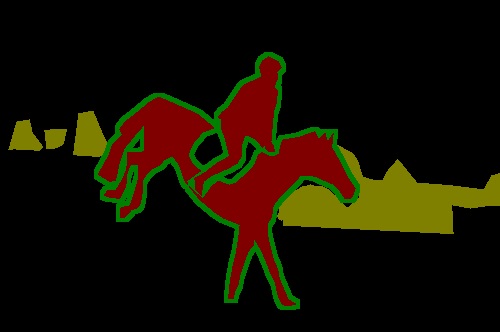}%
\label{p-3}} \hfil
\caption{Illustration of (a) a sample input image; (b) its original
labels from the PASCAL VOC dataset; and (c) its new labels with
maximum width $w=4$.}\label{objectboundary}
\end{figure}

A sample image, its
original and corresponding new labels are shown in Figs. \ref{objectboundary} (a), (b)
and (c), respectively. The white color in Fig. \ref{objectboundary} (b)
represents not only object boundaries, but also occluded objects in the
background that are difficult to recognize.  In contrast, the new label
system marks both the person and the horse in red, the thickened object
boundaries in green, the background in black, and the hard cases in
olive (or yellow green) in Fig. \ref{objectboundary} (c). We keep the
hard case region and its loss would be ignored during training as
in conventional methods. 

%%%%%%%%%%%%%%%%%%%%%%%%%%%%%%%%%%%%%%%%%%%%%%%%%%%%%%%%%%
%%%%%%%%%%%%%%%%%%%%%%%%%%%%%%%%%%%%%%%%%%%%%%%%%%%%%%%%%%

\subsection{Object Boundary Prediction FCN (OBP-FCN)}

\begin{figure}
\centering
\subfloat{\includegraphics[width=4.7in]{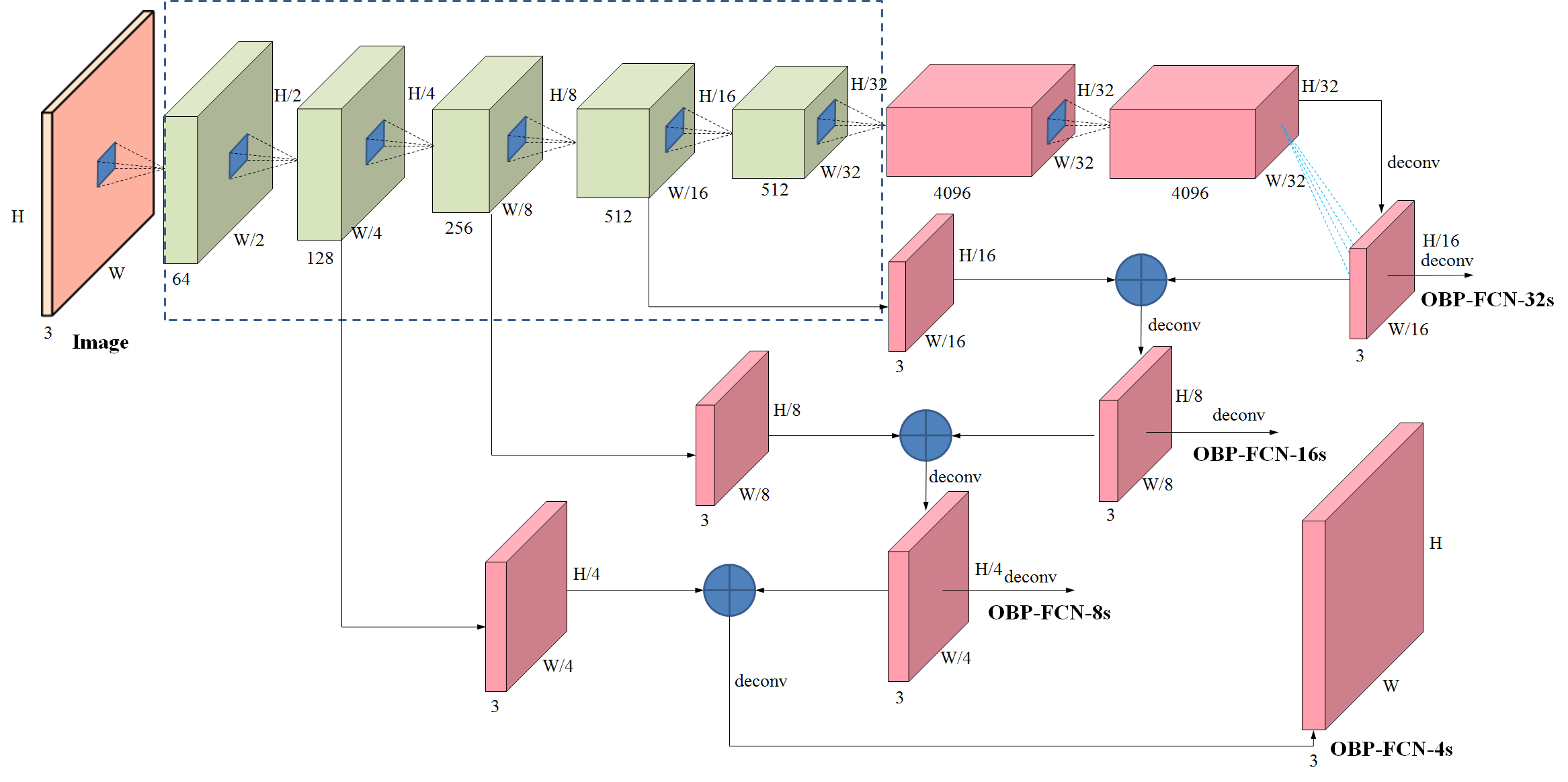}%
\label{p-1}} \hfil
\caption{The proposed OBP-FCN follows the basic structure of the FCN by
combining the coarse high level information with the detailed low level
information. Deconvolution is applied in upscaling. The response maps of
each output are summed element-wise by following a skip scheme.  All
network models (i.e., OBP-FCN-4s, OBP-FCN-8s, OBP-FCN-16s and
OBP-FCN-32s) and their results can be retrieved in each training step.
The final detail level is OBP-FCN-4s.}\label{OB_FCN}
\end{figure}

With the relabeled ground truth, we then train a network that can predict object (without class
distinction), boundary and background regions while ignoring the hard case
region. The network structure of the proposed network, called the
OBP-FCN, is shown in Fig.  \ref{OB_FCN}, where its first 5 layers have
the same convolution, pooling and ReLU operations as in VGG while the
fully connected layers 'fc-6' and 'fc-7' in VGG are replaced by two
convolutional layers.  

The unique characteristics of the OBP-FCN is that
it considers all features of 'pool4', 'pool3' and 'pool2' so as to combine large-scale 
class knowledge and detail boundary information. To initialize the
OBP-FCN, we begin with the VGG network \cite{simonyan2014very}
pre-trained on the ImageNet dataset \cite{deng2009imagenet}. Then, we
use the new 4-category labels to train the OBP-FCN for the desired goal. 

%%%%%%%%%%%%%%%%%%%%%%%%%%%%%%%%%%%%%%%%%%%%%%%%%%%%%%%%%%
%%%%%%%%%%%%%%%%%%%%%%%%%%%%%%%%%%%%%%%%%%%%%%%%%%%%%%%%%%

%Jay: Change OBP-FCN to OBP-FCN in the legend of the above figure.

We would like to elaborate the importance of thickened object boundaries
below. In the traditional FCN, the size of all kernels in VGG-16
convolution layers is $3 \times 3$. And with the help of the pooling layer, the gradually growing receptive
field of each layer allows the network to see patterns on different
scales.  The labeled object boundary
has an influence on learning local features, and it can force filters to
consider its existence at deeper layers. Without the constraint of labeled
object boundary, the original FCN network from \cite{long2015fully} stops at 
pool-3 as the performance does not improve furthermore. 

In contrast, we observe that the OBP-FCN continues to refine its object
boundary detection, benefiting from features in layer pool-2.  This is because 
the labeled maximum boundary width is two or four pixels, which can be seen on smaller scales. 

\section{Semantic Segmentation with OBG-FCN}\label{sec:OBG-FCN}

In this section, we propose an enhanced semantic segmentation solution,
called the object boundary guided FCN (OBG-FCN).  The object shape and
location information predicted by the OBP-FCN is used as a spatial mask
to guide the semantic segmentation task. An overview of the OBG-FCN is
given in Sec. \ref{subsec:overview}. One important subnet of the
OBG-FCN, called the OBG-Mask, is introduced in Sec. \ref{subsec:Mask}.
Some implementation details are discussed in Sec.
\ref{subsec:details}. 

\subsection{Overview of the OBG-FCN}\label{subsec:overview}

\begin{figure}
\centering
\subfloat{\includegraphics[width=4.7in]{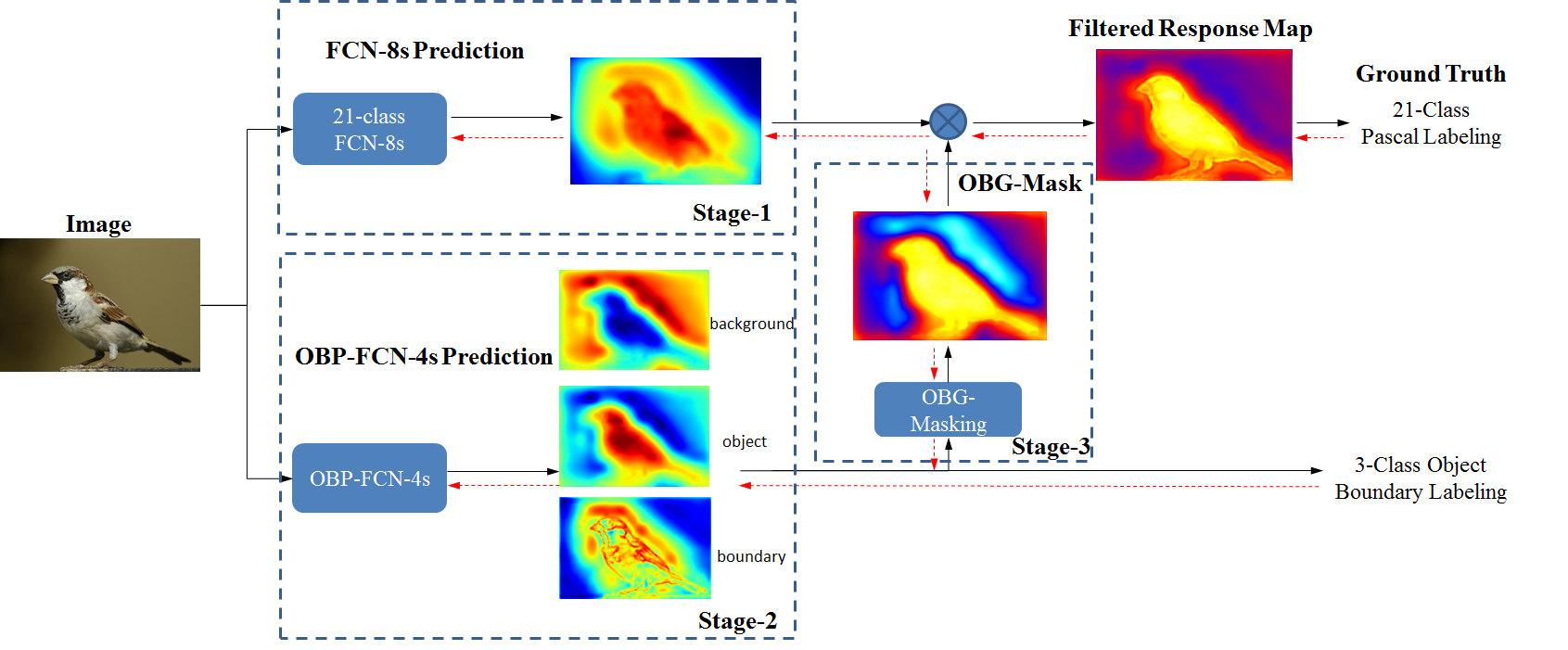}%
\label{p-1}} \hfil
\caption{Evolution of the response maps in the OBG-FCN: 1) the FCN-8s
provides a coarse class label for all objects (20 classes); 2) the
OBP-FCN-4s indicates the object localization without class distinction;
3) the OBG-Mask network produces an object mask, maps it to the
corresponding class label, and yield a more accurate filtered score map;
4) the final output of the OBG-FCN.}
\label{systemflow}
\end{figure}

The OBG-FCN system consists of three subnets; namely,
FCN-8s, OBP-FCN-4s and OBG-Mask. The evolution of the filter response
maps for an exemplary bird image is shown in Fig. \ref{systemflow}.
Since the output of the OBP-FCN is a 3-category map, we design a masking architecture 
called the OBG-Mask that passes the trained object shape and
localization information to the FCN-8s that segments 21 semantic classes (namely, 20 object
classes plus the background). This combination of two branches yields the final 21-class segmentation results.  We show the response score maps of
three subnets as well as the final output of the OBG-FCN in Fig.
\ref{systemflow}, which demonstrates the performance has improved significantly 
by integrating the three subnets.

%%%%%%%%%%%%%%%%%%%%%%%%%%%%%%%%%%%%%%%%%%%%%%%%%%%%%%%%%%
%%%%%%%%%%%%%%%%%%%%%%%%%%%%%%%%%%%%%%%%%%%%%%%%%%%%%%%%%%

%Jay: I am confused by your terminology of 20 and 21 classes.
%     The 20 classes are object classes. They do not include the background. Right?
%     What do you mean by 21 classes? I guess that it is 20 object classes plus the boundary.
%     Hope that my understanding is correct.

Multi-task learning is popular in recent CNN-based segmentation methods,
where extracted features are shared among multiple tasks.  However, when
we attempt to share features for the FCN-8s and the OBP-FCN, the
training tends to be biased on the FCN-8s sub-branch, resulting in poor performance 
of the OBP-FCN. For this reason, we train the OBP-FCN
separately and adopt the learned filter weights in the OBG-FCN
system afterwards. 

\subsection{OBG-Mask}\label{subsec:Mask}

The main purpose of the OBG-Mask subnet structure is to pass the 3-category
(object, boundary and background) labels obtained from the OBP-FCN to the
output of the FCN-8s to yield the ultimate output of the whole OBG-FCN
system. Specifically, the OBG-Mask subnet first converts the 3-category
inference result into 21-class object masks.  Then, the mask-level object
proposals are combined with the output of FCN-8s via element-wise
production. 

\begin{figure}
\centering
\subfloat{\includegraphics[width=2.8in]{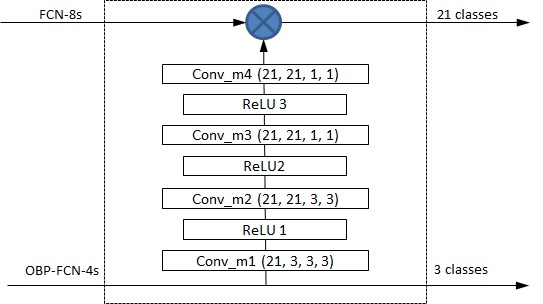}%
\label{p-1}} \hfil
\caption{An example of the OBG-Mask that accepts the output from the
OBP-FCN with $w=2$. The first two cascaded convolution layers have a
compound receptive field of $5 \times 5$, which is large enough to see
two adjacent boundaries. The masking architecture, then accepts the output from the
FCN-8s and conducts element-wise multiplication to
produce the final 21-class response score map.}\label{masking}
\end{figure}

One example of the OBG-Mask is shown in Fig. \ref{masking}, which
consists of three convolution and rectifier linear unit (ReLU) layers in
pair plus the 4th convolution layer. The last convolution layer,
conv-m4, does not include the ReLU layer. Its output is integrated with
the output from FCN-8s using either element-wise multiplication or summation,
which will be further discussed in the experiment section.
The parameters of each conv layer are given in the bracket, indicating the number of
filters, the number of input channels, the kernel height and width,
respectively. 

%%%%%%%%%%%%%%%%%%%%%%%%%%%%%%%%%%%%%%%%%%%%%%%%%%%%%%%%%%
%%%%%%%%%%%%%%%%%%%%%%%%%%%%%%%%%%%%%%%%%%%%%%%%%%%%%%%%%%

The maximum boundary width $w$ in the OBP-FCN determines the kernel size
of the convolutional layers. For example, the OBG-Mask shown in Fig.
\ref{masking}, is designed for the case of $w=2$.  The first two
cascaded convolution layers have a compound receptive field of $5 \times
5$, where two adjacent boundaries can be covered at the same time.
 In this way, the detected boundary can help
improve object or background labeling based on the local region
information and back-propagated class information. This is especially
beneficial to small objects and complex regions.  Although we can
benefit from a larger receptive field by increasing the kernal size,
this increases training complexity as well.  The proposed simple
structure is already sufficient to meet our needs. 

\subsection{Implementation Details}\label{subsec:details}

The full OBG-FCN system is designed to integrate the strengths of the
FCN-8s and the OBP-FCN. We first discuss its training procedure.  Both
filter weights of FCN-8s and OBP-FCN subnets are pre-trained and their
pre-trained values are used for initialization. For filter weights of
the OBG-Mask subnet, we adopt a random initialization scheme.  

The performance of the OBG-FCN is sensitive to its learning rate. We
adopt three training rates: 1) $10^{-15}$ for the FCN-8s; 2)$10^{-17}$
for the OBP-FCN; and 3) $10^{-10}$ for the OBG-Mark. The training rate
of $10^{-15}$ is commonly used for the FCN training. We adopt a lower
training rate for the OBP-FCN to ensure the provided object boundary
information is consistent and stable. We adopt a higher training rate
for the OBG-Mask so that its weights can be adjusted more aggressively
for faster converging. 

We trained multiple OBP-FCNs step-by-step: OBP-FCN-32s first,
OBP-FCN-16s next, OBP-FCN-8s afterwards, and OBP-FCN-4s last.  The
network learning rates are fixed at $10^{-10}$, $10^{-13}$, $10^{-14}$
and $10^{-15}$, respectively. The momentum is set to 0.99 as we use the full
image for training with a batch size of 1.  For the
training of OBP-FCN-32s, the weights of the first 5 convolutional layers
are copied from VGG and network surgery is performed to transform the
parameters of the original fully-connected form into the fully
convolutional form. 

The weight decay is set to 0.005 for OBP-FCN and 0.016 for OBG-FCN 
following the set-up of FCN and CRF-RNN, respectively.  We use the
standard softmax loss function, which is referred to as the
log-likelihood error function in \cite{krahenbuhl2013parameter}, and the
ReLU throughout the system for non-linearity. 

We implement the OBG-FCN system using the Caffe \cite{jia2014caffe}
library. The complete source code and trained models will be available
to the public.  Each training stage was conducted on a Titan-X graphic
card, and the training time varies depending on the number of training
images. 

\section{Experiments}\label{sec:exp}

In this section, we evaluate the performance of the proposed OBG-FCN on
the PASCAL VOC dataset.  We first evaluate the contributions of edge
labeling on object inference with the PASCAL VOC 11 dataset.   We then follow
\cite{zheng2015conditional} to train the proposed OBG-FCN framework with both
PASCAL VOC 2012 training image and an augmented PASCAL labeling
\cite{hariharan2014simultaneous}. The performance is evaluated on a
non-overlapping subset of the PASCAL VOC 2012 validation image set.
Finally, we compare the performance on both the PASCAL 2011 and 2012
test sets by submitting the proposed solution to the evaluation server. 

\subsection{Performance of OBP-FCN}\label{subsec:boundary}

We first evaluate the impact of object boundary labeling on the accuracy
of object region prediction. As mentioned in Sec. \ref{labelob}, we can
choose any desired maximum width in object boundary relabeling.  We
conduct experiments with three different maximum boundary widths and
compare the performance on the PASCAL VOC 2011. 1112 
images are used for the training of OBP-FCN, and 1111 images are used for validation. 
In order to compare with the original 20 object-class labels, we
retrain the original FCN model with 20 object classes using the training
data of the PASCAL VOC 2011. Then, we convert the segmentation result
into object labels without class distinction for fair comparison. 

We evaluate the performance on object region prediction by calculating
the accuracy between the predicted object area and the relabeled ground
truth.  By following \cite{long2015fully}, four evaluation metrics are
used, including pixel accuracy, mean accuracy, mean IU (Intersection over Union) and frequency
weighted IU.  The relabeled object boundaries are proven to be more effective
in predicting objects' detail shapes than the 20-class labeling.

%Jay: Please spell out IU when it is mentioned for the first time.

%===========================================================
\begin{table*}[]
\centering
\caption{Performance on object area prediction with different labeling methods.}
\label{table1}
\resizebox{\textwidth}{!}{%
\begin{tabular}{c|cccc|cccc|cccc|cccc}
 & \multicolumn{4}{c|}{pixel acc.} & \multicolumn{4}{c|}{mean acc.} & \multicolumn{4}{c|}{mean IU} & \multicolumn{4}{c}{f.w. IU} \\ \cline{2-17} 
 & 32s & 16s & 8s & 4s & 32s & 16s & 8s & 4s & 32s & 16s & 8s & 4s & 32s & 16s & 8s & 4s \\ \hline
FCN & 90.4 & 90.7 & 90.9 & 90.3 & 83.6 & 83.9 & 84.0 & 82.6 & 76.8 & 77.4 & 77.7 & 76.2 & 82.4 & 82.8 & 83.1 & 82.0 \\
\begin{tabular}[c]{@{}c@{}}0-pixel\\ OBP-FCN\end{tabular} & 90.7 & 91.2 & 91.3 & 91.0 & 84.8 & 86.1 & 86.4 & 85.5 & 77.8 & 79.1 & 79.3 & 78.5 & 82.9 & 83.9 & 84.0 & 83.5 \\
\begin{tabular}[c]{@{}c@{}}2-pixel\\ OBP-FCN\end{tabular} & \textbf{91.4} & \textbf{91.9} & \textbf{92.0} & \textbf{92.0} & \textbf{89.3} & \textbf{89.0} & \textbf{88.8} & \textbf{89.0} & \textbf{80.5} & \textbf{81.1} & \textbf{81.2} & \textbf{81.3} & \textbf{84.6} & \textbf{85.2} & \textbf{85.3} & \textbf{85.4} \\
\begin{tabular}[c]{@{}c@{}}4-pixel\\ OBP-FCN\end{tabular} & 91.1 & 91.3 & 91.5 & 91.4 & 86.9 & 87.1 & 87.1 & 87.2 & 79.2 & 79.6 & 79.9 & 79.8 & 83.7 & 84.1 & 84.4 & 84.3
\end{tabular}}
\end{table*}
%===========================================================

%%%%%%%%%%%%%%%%%%%%%%%%%%%%%%%%%%%%%%%%%%%%%%%%%%%%%%%%%%
\begin{figure}[t]
\captionsetup[subfigure]{labelformat=empty, position=top}
\centering
\subfloat[Input Image]{\includegraphics[width=0.7in]{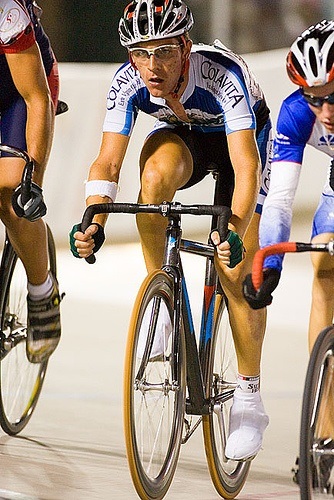}%
\label{p-1}} \hfil
\subfloat[FCN-8s]{\includegraphics[width=0.7in]{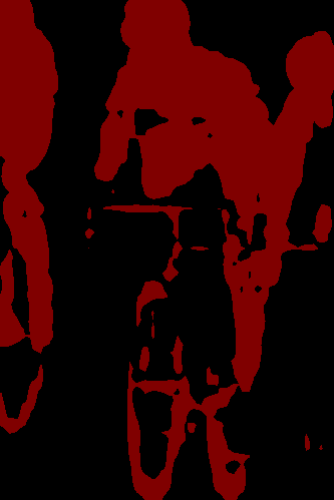}%
\label{p-1}} \hfil
\subfloat[OBP-FCN-4s (w=0)]{\includegraphics[width=0.7in]{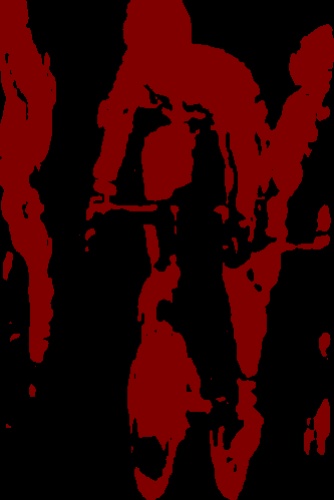}%
\label{p-2}}\hfil
\subfloat[OBP-FCN-4s (w=2)]{\includegraphics[width=0.7in]{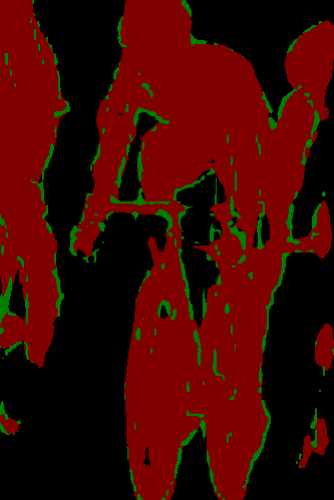}%
\label{p-3}} \hfil
\subfloat[OBP-FCN-4s (w=4)]{\includegraphics[width=0.7in]{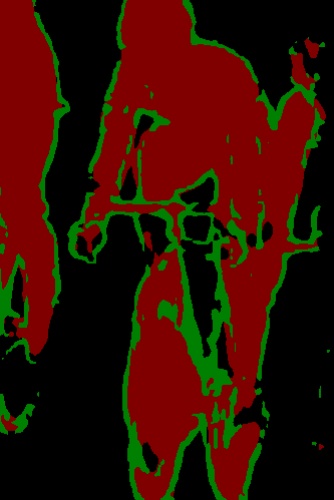}%
\label{p-4}} \hfil
\subfloat[Ground Truth]{\includegraphics[width=0.7in]{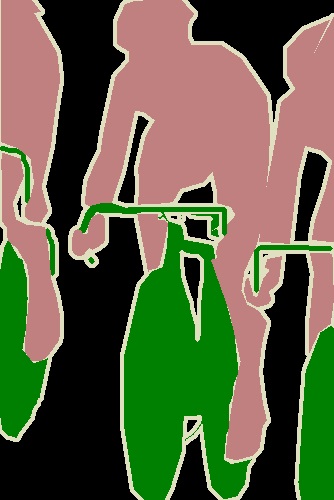}%
\label{p-4}} \hfil
\caption{Performance comparison of FCN-8s and OBP-FCN-4s with three 
maximum width values.}\label{edgemap}
\end{figure}
%%%%%%%%%%%%%%%%%%%%%%%%%%%%%%%%%%%%%%%%%%%%%%%%%%%%%%%%%%

%%%%%%%%%%%%%%%%%%%%%%%%%%%%%%%%%%%%%%%%%%%%%%%%%%%%%%%%%%
\begin{figure}[htb]
\captionsetup[subfigure]{labelformat=empty,position=top}
\centering
\subfloat[OBP-FCN-32s]{\includegraphics[width=0.82in, height=1.2in]{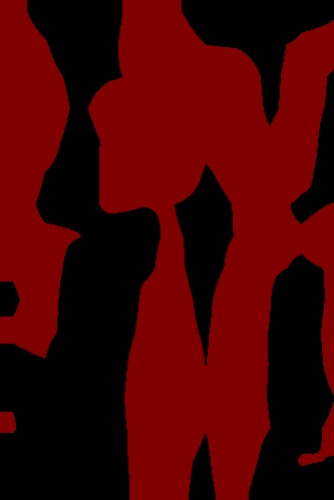}%
\label{p-2}}\hfil
\subfloat[OBP-FCN-16s]{\includegraphics[width=0.82in, height=1.2in]{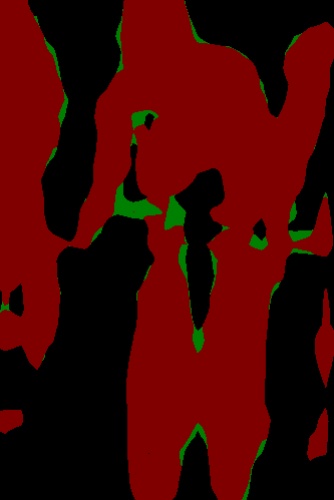}%
\label{p-3}} \hfil
\subfloat[OBP-FCN-8s]{\includegraphics[width=0.82in, height=1.2in]{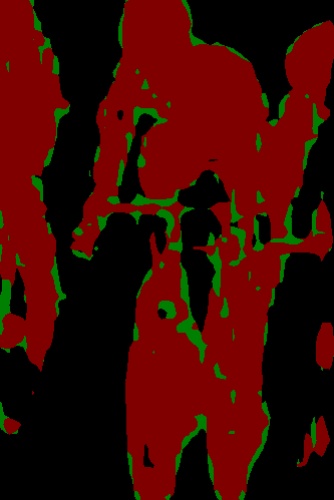}%
\label{p-4}} \hfil
\subfloat[OBP-FCN-4s]{\includegraphics[width=0.82in, height=1.2in]{2ob.png}%
\label{p-4}} \hfil
\subfloat[New Label]{\includegraphics[width=0.82in, height=1.2in]{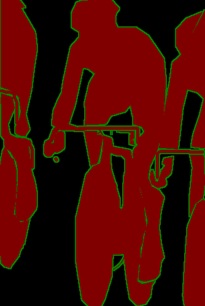}%
\label{p-4}} \hfil
\caption{Performance comparison of four OBP networks with maximum
width $w=2$: OBP-FCN-32s, OBP-FCN-16s, OBP-FCN-8s and OBP-FCN-4s.}\label{revomap}
\end{figure}
%%%%%%%%%%%%%%%%%%%%%%%%%%%%%%%%%%%%%%%%%%%%%%%%%%%%%%%%%%

We compare object shape prediction results of FCN-8s and OBP-FCN-4s with
three maximum width values ($w=0$, $2$ and $4$.) on a test image
in Fig. \ref{edgemap}, where the original input and the
ground truth label are also provided. We see that object boundary
relabeling and training does help improve object localization in
providing more object details and avoiding false alarms. The case $w=2$
gives the best performance. It is worthwhile to point out that we do not
evaluate the performance of the object boundaries prediction since we only
use boundaries in the training to help object/background or
object/object segmentation.  Thus, even if the predicted object
boundary (in green color) may not be closed, it still contributes to the
completeness of object prediction when all results are integrated in a
later stage. 

In addition, the results indicate that with $w=2$, the
accuracy of object inference can be further improved even on
OBP-FCN-4s. In Fig. \ref{revomap}, we show that by gradually combining the
low-level features, the object contours begin to emerge and OBP-FCN-4s gives 
the best result with natural  and sufficient details. 

\subsection{Performance of OBG-FCN}\label{subsec:performance-obj}

As described in Sec. \ref{sec:OBP-FCN}, we train the proposed OBP-FCN-4s
with 11,685 relabeled ground truth images, including 1,464 labeled
images in the PASCAL VOC 2012 trainging set and the augmented labeled
images from \cite{hariharan2014simultaneous}. Then, we adopt the
pre-trained FCN-8s model from \cite{long2015fully} and conduct the
end-to-end training to get the final OBG-FCN.  Since the labeling of
augmented data is not very accurate, we use the full set of 11,685
images to train the OBP-FCN-32s only. This is done because a large
amount of image data can provide rich yet coarse information of object
features. Then, we train the remaining two subnets of the OBG-FCN with
the 1,464 labeled images in the PASCAL VOC 2012 datset. The accurately
labeled object boundaries help construct more accurate object shapes, leading to better segmentation results. 

\noindent
{\bf Performance on Validation Set}  Since there is an overlap between the augmented labeled image set and
the PASCAL VOC 12 validation set, we select a list of 346
non-overlapping images from the PASCAL VOC 12 validation set, and
evaluate the performance of the proposed OBG-FCN on this subset. The results of the baseline FCN-8s, along with the proposed OBG-FCN with relabeled boundaries of 2-pixel
and 4-pixel maximum widths are presented in Table \ref{table2}.  
We see that the OBG-FCN with $w=2$ offers the best results. 

%%%%%%%%%%%%%%%%%%%%%%%%%%%%%%%%%%%%%%%%%%%%%%%%%%%%%%%%%%
\begin{figure}[htb]
\captionsetup[subfigure]{labelformat=empty,position=top}
\centering
\subfloat[Input Image]{\includegraphics[width=0.9in, height=0.6in]{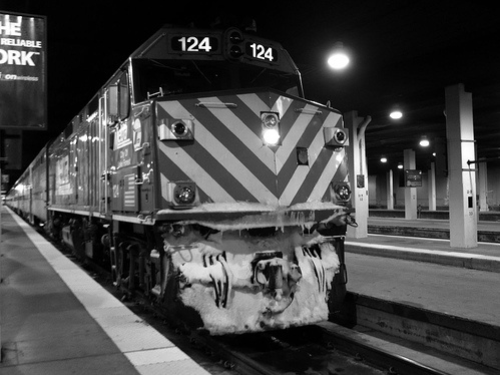}%
\label{p-1}} \hfil
\subfloat[FCN-8s]{\includegraphics[width=0.9in, height=0.6in]{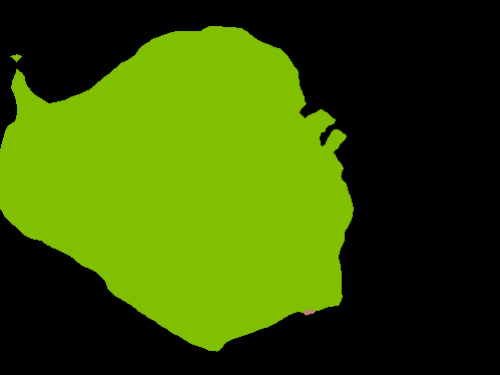}%
\label{p-2}}\hfil
\subfloat[OBP-FCN-4s]{\includegraphics[width=0.9in, height=0.6in]{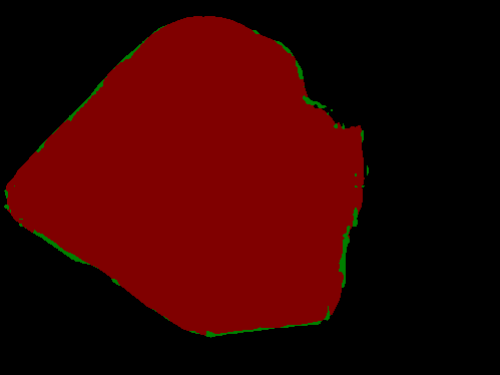}%
\label{p-3}} \hfil
\subfloat[OBG-FCN]{\includegraphics[width=0.9in, height=0.6in]{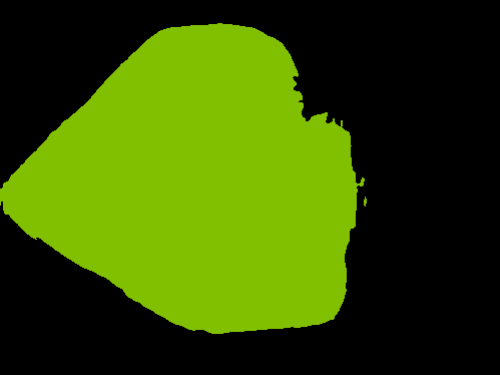}%
\label{p-4}} \hfil
\subfloat[Ground Truth]{\includegraphics[width=0.9in, height=0.6in]{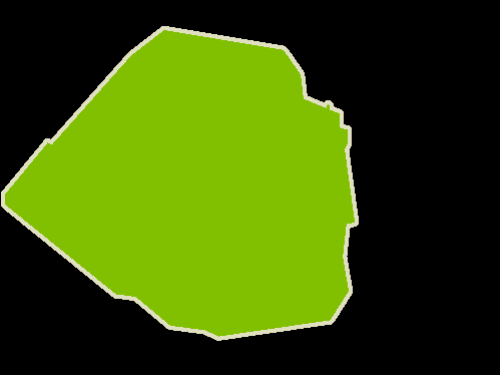}%
\label{p-4}} \hfil
\subfloat{\includegraphics[width=0.9in]{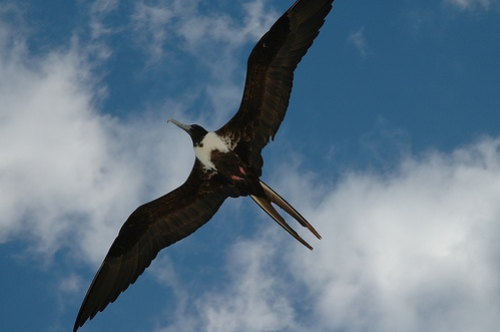}%
\label{p-1}} \hfil
\subfloat{\includegraphics[width=0.9in]{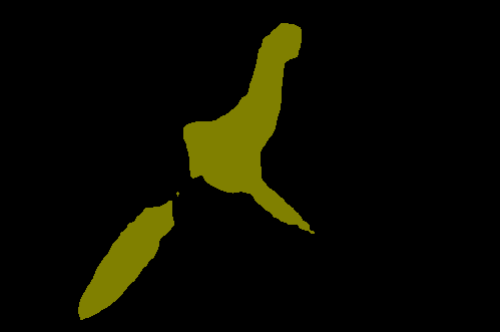}%
\label{p-2}}\hfil
\subfloat{\includegraphics[width=0.9in]{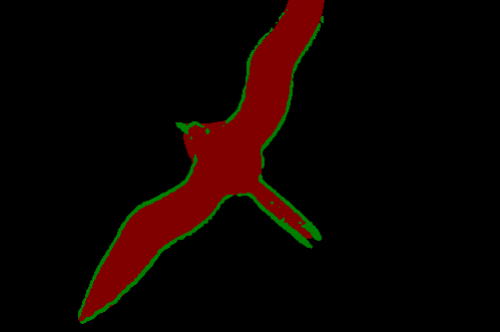}%
\label{p-3}} \hfil
\subfloat{\includegraphics[width=0.9in ]{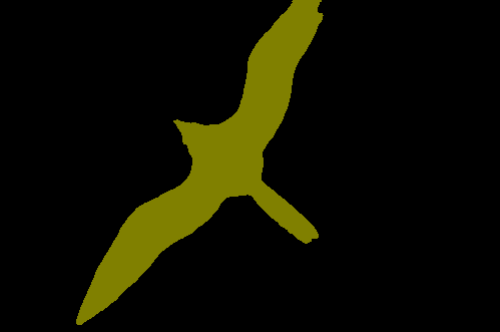}%
\label{p-4}} \hfil
\subfloat{\includegraphics[width=0.9in ]{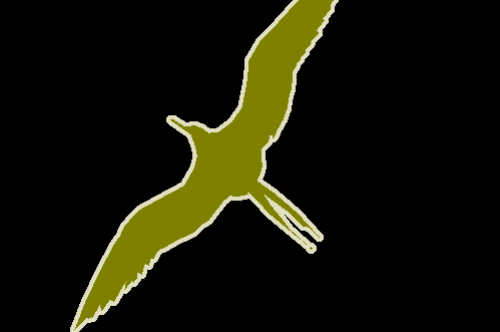}%
\label{p-4}} \hfil
\subfloat{\includegraphics[width=0.9in, height=1.0in]{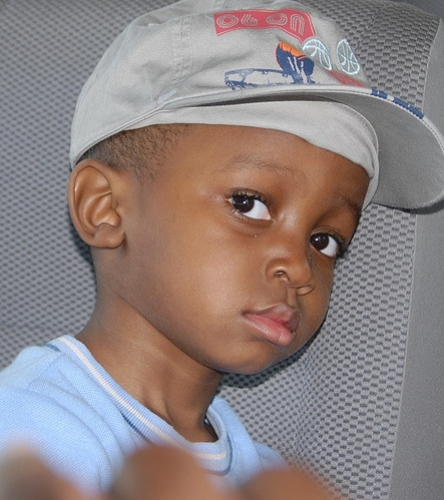}%
\label{p-1}} \hfil
\subfloat{\includegraphics[width=0.9in, height=1.0in]{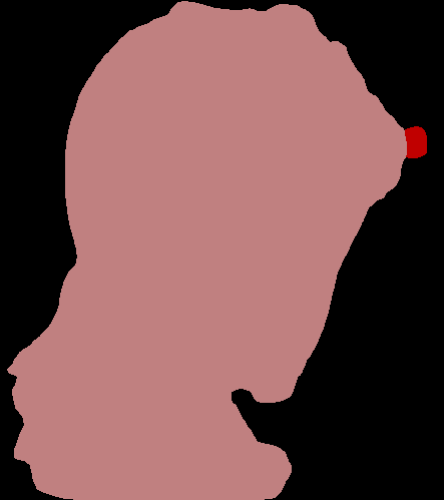}%
\label{p-2}}\hfil
\subfloat{\includegraphics[width=0.9in, height=1.0in]{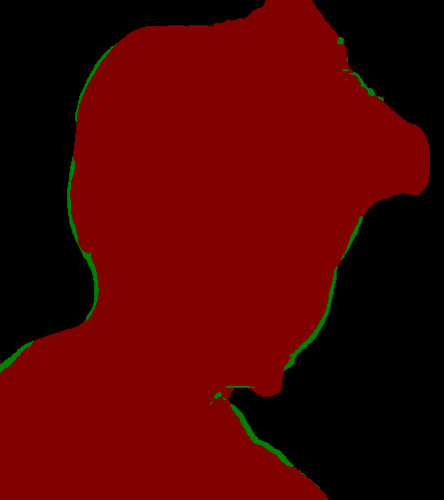}%
\label{p-3}} \hfil
\subfloat{\includegraphics[width=0.9in, height=1.0in]{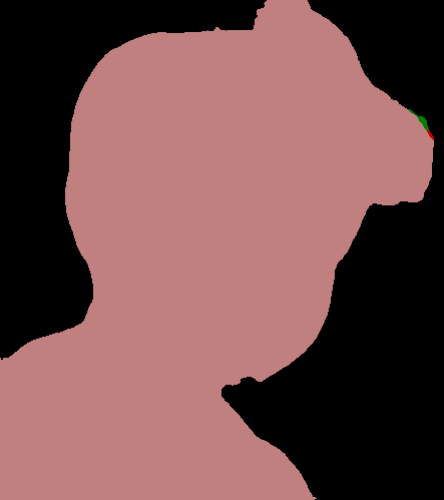}%
\label{p-4}} \hfil
\subfloat{\includegraphics[width=0.9in, height=1.0in]{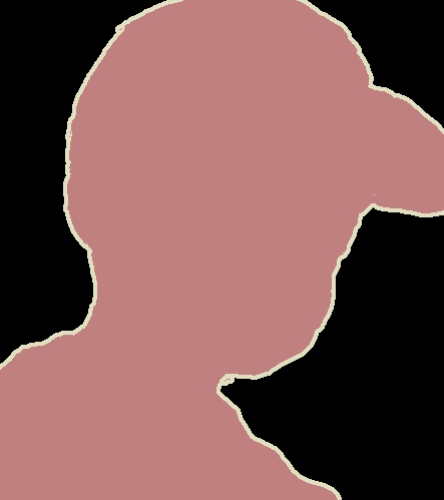}%
\label{p-4}} \hfil
\subfloat{\includegraphics[width=0.9in, height=1.1in]{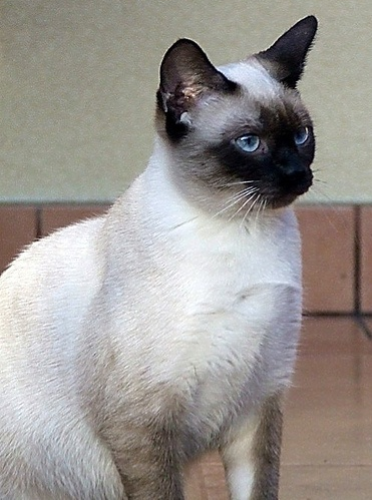}%
\label{p-1}} \hfil
\subfloat{\includegraphics[width=0.9in, height=1.1in]{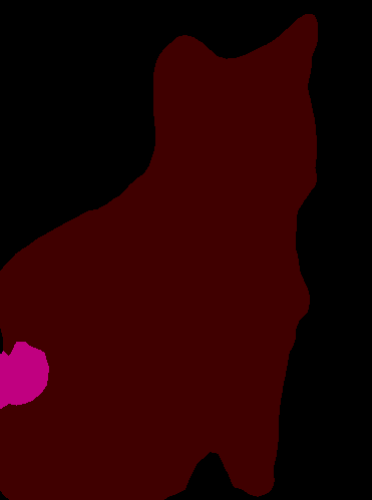}%
\label{p-2}}\hfil
\subfloat{\includegraphics[width=0.9in, height=1.1in]{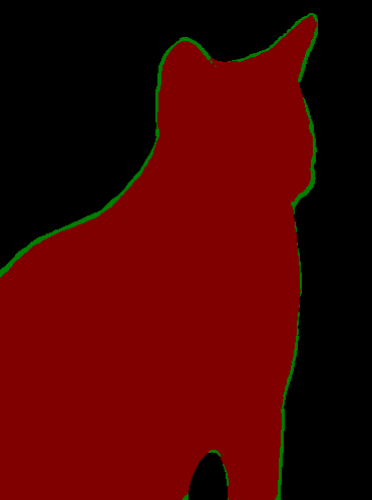}%
\label{p-3}} \hfil
\subfloat{\includegraphics[width=0.9in, height=1.1in ]{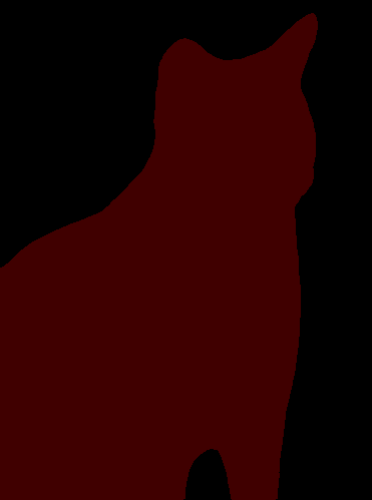}%
\label{p-4}} \hfil
\subfloat{\includegraphics[width=0.9in, height=1.1in]{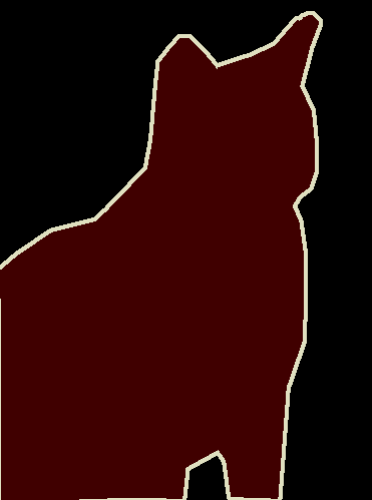}%
\label{p-4}} \hfil
\caption{Visualization of exemplary single-class segmentation results in VOC2012 validation set
(from left to right and best viewed in color): input images,
intermediate results of FCN-8s, intermediate results of OBP-FCN-4s, and
final results of OBG-FCN, and the ground truth labels.}\label{result1}
\end{figure}
%%%%%%%%%%%%%%%%%%%%%%%%%%%%%%%%%%%%%%%%%%%%%%%%%%%%%%%%%%

%===========================================================
\begin{table}[h]
\centering
\caption{Performance on selective PASCAL VOC 2012 validation set of OBG-FCN with different
labeling pixel widths (w).}
\label{table2}
\begin{tabular}{c|c|c|c|c}
 & pixel acc. & mean acc. & mean IU & f.w. IU \\ \hline
FCN-8s & 90.1 & 74.1 & 61.1 & 82.7 \\
OBG-FCN ($w=2$) & \textbf{91.6} & \textbf{76.4} & \textbf{64.9} & \textbf{85.3} \\
OBG-FCN ($w=4$) & 91.5 & 75.5 & 64.5 & 84.9
\end{tabular}
\end{table}
%===========================================================

Exemplary segmentation results of single-class images are shown in Fig.
\ref{result1}. Compared with FCN-8s, more accurate silhouettes are obtained by the OBP-FCN-4s
since the relabeled boundary offers extra information to constraint the object. As a
result, the OBG-FCN can improve FCN's results by either providing some
lost object information (e.g. bird's wing) or correcting some false
decisions (e.g. train's tail). The boundary regions are smoother and
more natural. 

\begin{figure}[t]
\captionsetup[subfigure]{labelformat=empty,position=top}
\centering
\subfloat[Input Image]{\includegraphics[width=0.9in]{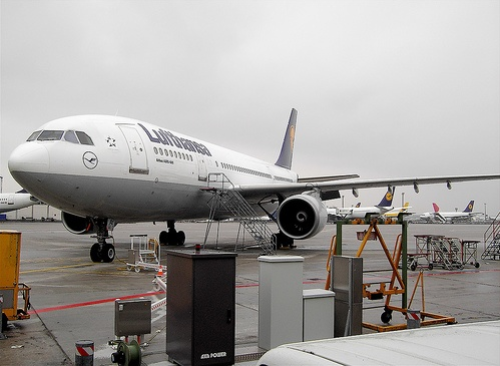}%
\label{p-1}} \hfil
\subfloat[FCN-8s]{\includegraphics[width=0.9in]{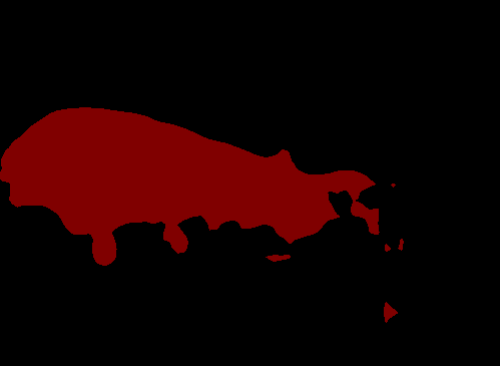}%
\label{p-2}}\hfil
\subfloat[OBP-FCN]{\includegraphics[width=0.9in]{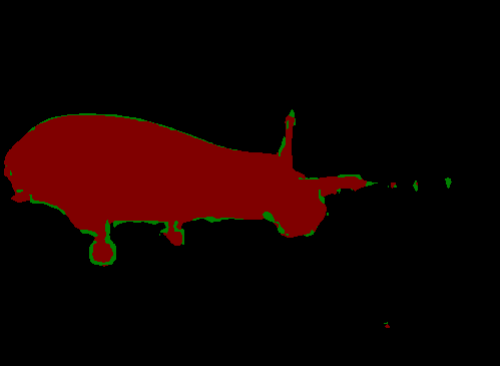}%
\label{p-3}} \hfil
\subfloat[OBG-FCN]{\includegraphics[width=0.9in ]{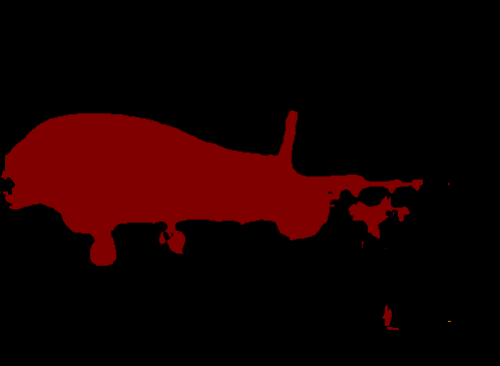}%
\label{p-4}} \hfil
\subfloat[Ground Truth]{\includegraphics[width=0.9in ]{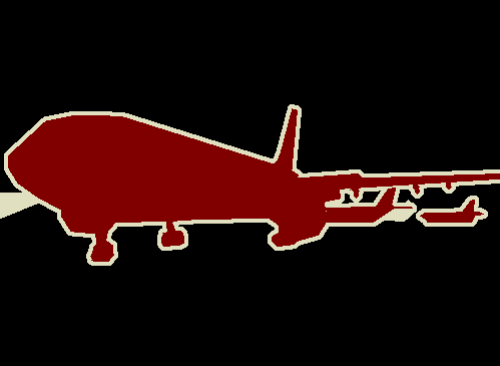}%
\label{p-4}} \hfil
\subfloat{\includegraphics[width=0.9in]{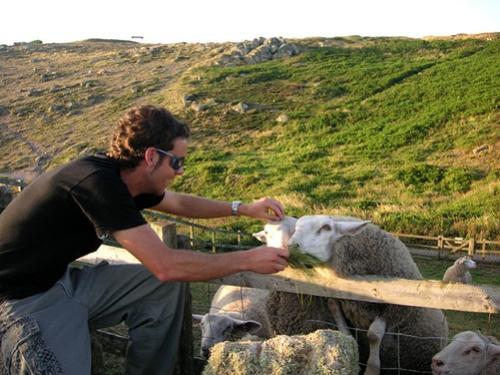}%
\label{p-1}} \hfil
\subfloat{\includegraphics[width=0.9in]{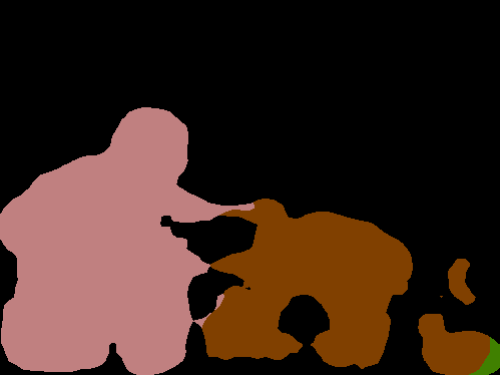}%
\label{p-2}}\hfil
\subfloat{\includegraphics[width=0.9in]{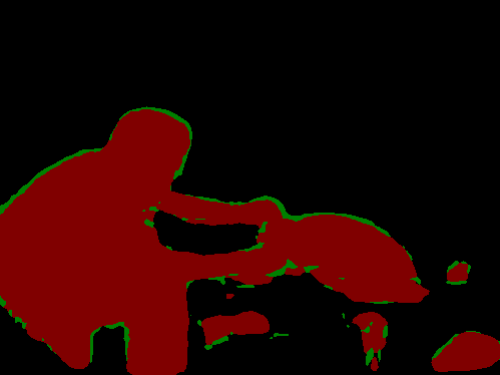}%
\label{p-3}} \hfil
\subfloat{\includegraphics[width=0.9in ]{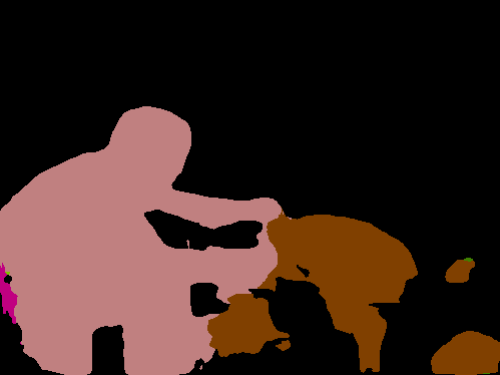}%
\label{p-4}} \hfil
\subfloat{\includegraphics[width=0.9in ]{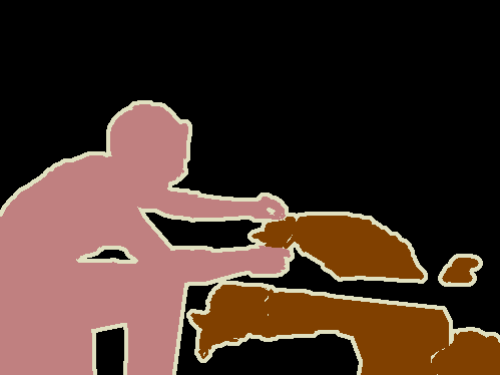}%
\label{p-4}} \hfil
\subfloat{\includegraphics[width=0.9in]{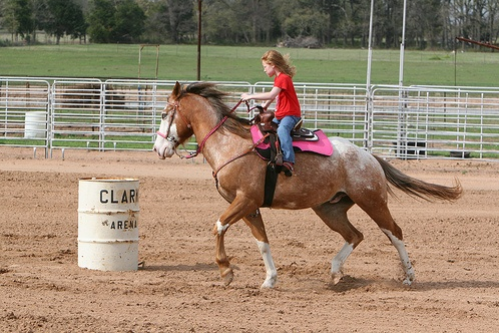}%
\label{p-1}} \hfil
\subfloat{\includegraphics[width=0.9in]{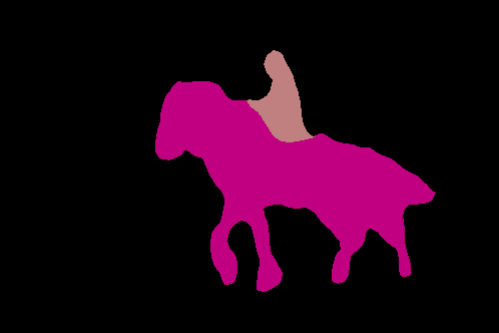}%
\label{p-2}}\hfil
\subfloat{\includegraphics[width=0.9in]{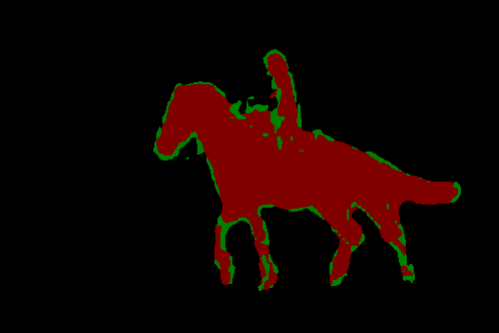}%
\label{p-3}} \hfil
\subfloat{\includegraphics[width=0.9in ]{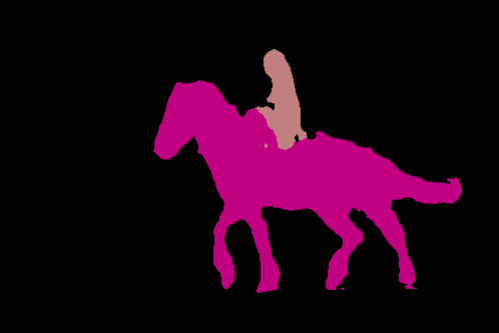}%
\label{p-4}} \hfil
\subfloat{\includegraphics[width=0.9in ]{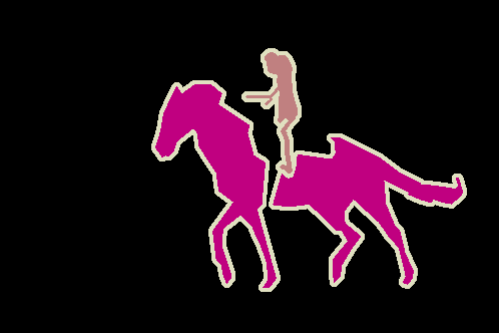}%
\label{p-4}} \hfil
\subfloat{\includegraphics[width=0.9in]{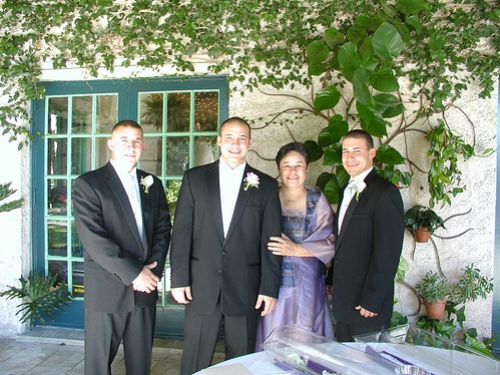}%
\label{p-1}} \hfil
\subfloat{\includegraphics[width=0.9in]{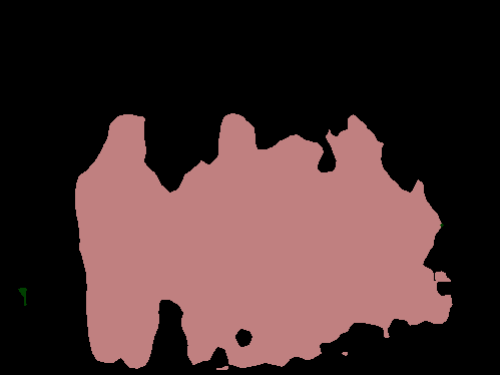}%
\label{p-2}}\hfil
\subfloat{\includegraphics[width=0.9in]{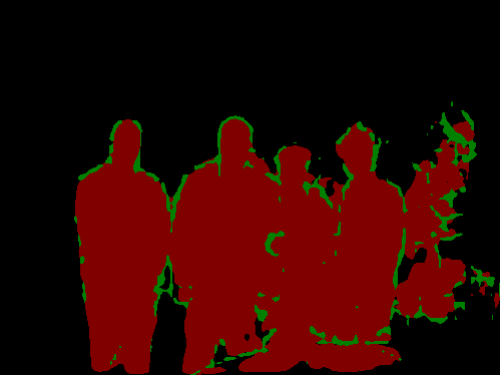}%
\label{p-3}} \hfil
\subfloat{\includegraphics[width=0.9in ]{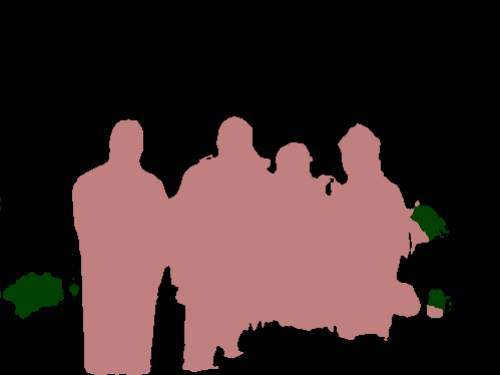}%
\label{p-4}} \hfil
\subfloat{\includegraphics[width=0.9in ]{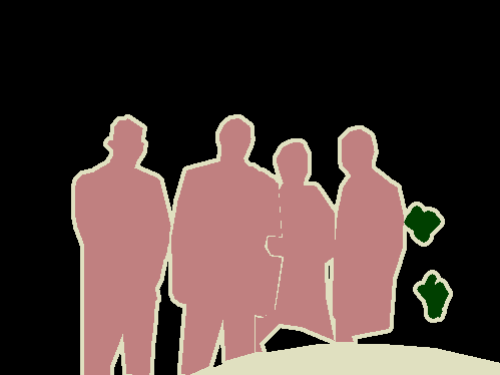}%
\label{p-4}} \hfil

\caption{Examples of multi-class and multi-object segmentation results in VOC2012 validation set.
Best viewed in color.}\label{result2}
\end{figure}

Furthermore, we compare the segmentation results of FCN-8s and OBG-FCN
for several exemplary multi-class and multi-object images in Fig.
\ref{result2}. We see that the proposed OBP-FCN-4s can
localize some boundaries between concatenated objects (e.g.  boundaries
between humans in the last row). Generally speaking, the segmentation
results of OBG-FCN are better than those of FCN-8s. 

\noindent
{\bf Comparison of Different Structure} As mentioned in Sec. \ref{subsec:Mask}, the combination of two
fully convolutional branches can be either element-wise multiplication or summation. Therefore, we
evaluate the performance of these two different settings to show that they bring with similar results. 
Furthermore, we present another set of results using only OBP-FCN-8s in stage 2 to construct
the OBG-FCN system. Results in Table. \ref{table3} prove that OBP-FCN-4s does provide a better benchmark result
with refined object details.

%%%%%%%%%%%%%%%%%%%%%%%%%%%%%%%%%%%%%%%%%%%%%%%%%%%%%%%%%%
%%%%%%%%%%%%%%%%%%%%%%%%%%%%%%%%%%%%%%%%%%%%%%%%%%%%%%%%%%
%===========================================================
\begin{table}[h]
\centering
\caption{Comparison of mean IU performance using multiplication or
summation for subnet fusion and using the OBP-FCN-4s or the OBP-FCN-8s.}
\label{table3}
\begin{tabular}{c|c|l|c|l}
\multirow{2}{*}{\begin{tabular}[c]{@{}c@{}}OBG-Mask \\ method\end{tabular}} & \multicolumn{2}{c|}{Product} & \multicolumn{2}{c}{Summation} \\ \cline{2-5} 
 & \multicolumn{1}{l|}{OBP-FCN-8s} & OBP-FCN-4s & \multicolumn{1}{l|}{OBP-FCN-8s} & OBP-FCN-4s \\ \hline
Mean IU & 64.5 & \multicolumn{1}{c|}{64.8} & 64.5 & \multicolumn{1}{c}{64.9}
\end{tabular}
\end{table}
%===========================================================

\noindent
{\bf Performance on PASCAL VOC Test Set} We then use the same training data set and evaluate the performance
on the PASCAL VOC 2011 and 2012 test sets by submitting the results to
the evaluation server. The results are given in Table \ref{table4}.  As
shown in this table, the OBG-FCN with $w=2$ reaches 69.5\% mean IU in VOC 2011 test and
69.1\% in VOC 2012 test, outperforming the baseline FCN
by about $7\%$. It has also surpassed the previous state-of-the-art methods without relying
on conditional random fields. 

%===========================================================
\begin{table}[h]
\centering
\caption{Performance comparison of nean IU accuracy on the PASCAL VOC 2011 and 2012 
test datasets between FCN-8s, DeepLab, DT-SE, DT-EdgeNet and OBG-FCN with $w=2$ and $w=4$.}
\label{table4}
\resizebox{\textwidth}{!}{%
\begin{tabular}{c|c|c|c|c|c|c}
 & FCN-8s \cite{long2015fully} & DeepLab \cite{chen2014semantic} & DT-SE \cite{chen2015semantic} & OBG-FCN (w=4) & DT-EdgeNet \cite{chen2015semantic}& OBG-FCN (w=2) \\ \hline
VOC2011 test & 62.7 & / & / & 68.9 & / & 69.5 \\
VOC2012 test & 62.2 & 65.1 & 67.8 & 68.6 & 69.0 & 69.1
\end{tabular}}
\end{table}
%===========================================================

\noindent
{\bf Comparison of Inference Speed} We finally evaluate the inference speed of the proposed framework. 
As in Table. \ref{table5}, the OBG-FCN takes about 0.187 s/image in average 
on a Titan X GPU, which offers possibilities for real-time segmentation. 
The inference time on the Intel Core i7-5930 CPU takes about 5.6 s/image in average.
Compared with FCN-8s, the proposed OBG-FCN takes about twice the inference time as we rely on
two distinct FCN sub-branches. We further test on a publicly available model of CRF-RNN 
with 10 mean-filed iterations. And although the CRF-RNN provides the best 72.0 $\%$ mean IU in VOC12 test set,
its average inference time on GPU is ten times slower
than the proposed OBG-FCN.
%===========================================================
\begin{table}[]
\centering
\caption{Average inference time (s/image).}
\label{table5}
\begin{tabular}{c|c|c|c}
 & FCN-8s & OBG-FCN & CRF-RNN \\ \hline
CPU time & 2.56 & 5.34 & 5.21 \\
GPU time & 0.09 & 0.186 & 1.92
\end{tabular}
\end{table}
%===========================================================

\section{Conclusion and Future Work}\label{sec:conclusion}

In this work, we propose a fully-convolutional network with two distinct branches in earlier stage to specifically learn 
the class information and the mask-level object proposal. The strengths of two sub-networks are then
fused together with a proposed OBG-Mask architecture to provide better semantic segmentation results. 
The method was proven to be effective in generating accurate object localizations and refined object details. 

Although the proposed OBP-FCN subnet can provide a better object shape
constraint and yield better semantic segmentation results, the final
performance is still limited by the accuracy of the 20-class FCN subnet.
It is important to find a better baseline in building the full OBG-FCN
system. Also, the performance can be further improved if better labeling
can be provided for more training images. Generally speaking, semantic
segmentation remains to be a challenging problem for further research.

%===========================================================
\bibliographystyle{splncs}
\bibliography{egbib}

%this would normally be the end of your paper, but you may also have an appendix
%within the given limit of number of pages
\end{document}